\documentclass[11pt]{article}
\usepackage{coling2016}
\usepackage{times}
\usepackage{url}
\usepackage{collref}
\usepackage{latexsym}
\usepackage{multirow}
\usepackage{amsmath}
\usepackage{caption}
\DeclareMathOperator*{\argmax}{argmax}

\title{Real Multi-Sense or Pseudo Multi-Sense: \\An Approach to Improve Word Representation}

\author{Haoyue Shi$^1$, Caihua Li$^1$ and Junfeng Hu$^{1,2*}$ \\
$^1$ School of Electronics Engineering and Computer Science, \\ Peking University, Beijing, China \\
$^2$ Key Laboratory of Computational Linguistics,\\ Ministry of Education, Peking University, Beijing, China\\
{\tt \{hyshi,peterli,hujf\}@pku.edu.cn}}
\date{}

\begin{document}

\maketitle
\begin{abstract}
Previous researches have shown that learning multiple representations for polysemous words can improve the performance of word embeddings on many tasks. However, this leads to another problem. Several vectors of a word may actually point to the same meaning, namely pseudo multi-sense. In this paper, we introduce the concept of pseudo multi-sense, and then propose an algorithm to detect such cases. With the consideration of the detected pseudo multi-sense cases, we try to refine the existing word embeddings to eliminate the influence of pseudo multi-sense. Moreover, we apply our algorithm on previous released multi-sense word embeddings and tested it on artificial word similarity tasks and the analogy task. The result of the experiments shows that diminishing pseudo multi-sense can improve the quality of word representations. Thus, our method is actually an efficient way to reduce linguistic complexity. 
\end{abstract}

\section{Introduction}
\blfootnote{
    This work is licenced under a Creative Commons 
    Attribution 4.0 International License.
    License details:
    \url{http://creativecommons.org/licenses/by/4.0/}
}
\blfootnote{
This work is supported by the National Natural Science Foundation of China (grant No.61472017, M1552004).
}

\par
Representing meanings of words by embedding them into a high dimensional vector space, so called word embedding, is a useful technique in natural language processing. An intuitive idea is to encode one word into a single vector, which contains the semantic information of the word in corpus \cite{bengio2003neural, collobert2008unified, mnih2007three, mikolov2010recurrent}.

\par
There is a consensus that natural languages always include lots of polysemous words. For example, when the word {\sl star} appears together with words like {\sl planet, satellite}, it may roughly denote a kind of celestial body; when {\sl star} appears with words like {\sl movie, song, drama}, it may stand for a famous person. For most cases, we human beings can easily point out which sense a word belongs to based on its context. Considering the polysemous words, some previous approaches have learned multiple embeddings for a word, discriminating different senses by their context, related syntax and topics \cite{reisinger2010multi, huang2012improving, chen2014unified, pina2014simple, neelakantan2015efficient, cheng2015syntax, liu2015topical}. The authors also provided methods to disambiguate among the multiple representations. \newcite{li2015multi} have demonstrated that multi-sense word embeddings could be helpful to improve the performance on many NLP and NLU tasks.
\par
However, this leads to another problem. It's much more difficult for computer than human beings to detect whether two appearances of a same word stand for the same sense. Moreover, the contexts may be totally different even if these appearances belong to the same meaning based on human judgement. Previous multi-sense word embedding approaches often tend to embed a word in such situation into more than one vector by mistake (actually, they have the same meaning and should be embedded into only one vector). Consider three different representations of word {\sl bear} learnt by the method introduced by \newcite{neelakantan2015efficient}, which are shown by their nearest neighbors in the vector space {\sl MSSG-50d}.
\begin{itemize}
\item emerald, \textbf{bears}, \textbf{three-toed}, \textbf{snake}, \textbf{periwinkle}, \textbf{ruffed}, \textbf{hoopoe}, distinctive, unmistakable
\item \textbf{bird}, \textbf{wolf}, arrow, \textbf{pelican}, emerald, canyon, diamond, \textbf{buck}, \textbf{deer}
\item pride, lady, hide, king, gift, crane, afflict, promise, reap, protect
\end{itemize}
The words clearly related to the domain {\sl animals} are bolded. We could infer that the first two representations have the same meaning that points to the animal bear, and the third representation has different meaning. We call such different learnt representations of a word with the same meaning (e.g. the first two representations of word {\sl bear} shown above) {\sl pseudo multi-sense}, where we judge whether senses are pseudo multi-sense by comparing their domains.
\par
Given the word embeddings, which have multiple vectors for each polysemous word, we introduce an algorithm based on domains and semantic relations to detect pseudo multi-sense, since word representations which stand for the same meaning would have the same hypernym and belong to the same domain. Then we try to eliminate the effect of pseudo multi-sense by training a global transition matrix which projects the original word vectors into a new vector space based on the detected pseudo multi-sense pairs, minimizing the distance between pseudo multi-sense pairs in the vector space while keeping the spatial relation of other pairs. We propose the algorithm in Section 3 and evaluate it in Section 4.
\par
Obviously, detecting and diminishing pseudo multi-sense would make word sense representations, which can be processed by computer, closer to human thinking. We also suggest this approach can improve the performance on real world NLU tasks by evaluating the algorithm on the analogy test dataset introduced by \newcite{mikolov2013efficient}, and also on WordSim-353 \cite{finkelstein2001placing} and SCWS \cite{huang2012improving} dataset which include human judgements on similarity between pairs of words.

\section{Background and related work}
\subsection{Distributional word representations}
Since \newcite{bengio2003neural} applied neural network to language model, which treats word embeddings as parameters and thus it allows us to learn the language model and word embeddings at the same time, many researchers have proposed other neural network models \cite{mnih2007three, collobert2008unified, mikolov2013efficient} to improve in both efficiency and accuracy. What's more, hierarchical softmax by \newcite{morin2005hierarchical}, noise contrastive estimation by \newcite{mnih2013learning} and negative sampling by \newcite{mikolov2013linguistic} make it possible to learn accurate word embeddings in a short time.
\subsection{Multi-sense word embeddings}
\par
Most vector-space models (VSMs) represent a word with only one vector, which clearly fails to capture homonymy and polysemy. And thus, \newcite{huang2012improving} proposed a method to generate the context embeddings in the following way. Firstly, they generate single-sense word embeddings and compute out the context embeddings. Then they cluster the context embeddings, and the result are used to re-label each occurrence of each word in the corpus. Thirdly, the model they proposed is applied to the labeled corpus to generate the multi-sense embeddings. \newcite{chen2014unified} took external knowledge base into consideration and built a model to learn a separate vector for each sense pre-defined by WordNet \cite{miller1995wordnet}. \newcite{neelakantan2015efficient} improved multi-sense word embedding model by dropping the assumption that each word should have the same number of senses, and proposed a  non-parametric model to automatically discover a varying number of senses per word type. \newcite{cheng2015syntax} proposed a syntax-aware approach for multi-sense word embeddings. 
\subsection{WordNet and WordNet domain knowledge}
\par
WordNet \cite{miller1995wordnet} is a large lexical database of English. Nouns, verbs, adjectives and adverbs are grouped into sets of cognitive synonyms, namely synsets, each expressing a distinct concept. Synsets are represented by a word, a pos tag and a label, and interlinked by means of conceptual-semantic and lexical relations (hypernymy/hyponymy).  \newcite{chen2014unified} used WordNet to improve word embeddings.
\par
\newcite{magnini2000integrating} and \newcite{bentivogli2004revising} presented a WordNet Domains Hierarchy, which is a language-independent resourse composed of 164 domain labels. What's more, \newcite{gonzalez2012graph} provided a graph based improvement and released a domain knowledge (Extended WordNet Domains) base aligned to WordNet 3.0, which we use in our experiments as domain knowledge. Extended WordNet Domains contains 170 domains and the probability of each synset in WordNet 3.0 in each domain. The domains it provided include {\sl acoustics, agriculture, volleyball, etc.}
\subsection{Vector space projection}
Even though bilingual data always plays an important role in the modern statistical machine translation system, it had failed to map the missing word and phrase entries between two languages until \newcite{mikolov2013exploiting} proposed a simple but effective method to extend dictionaries and translation tables. The main idea of this novel method is to learn a linear projection between the languages using a small bilingual dictionary but making little assumption about the languages, which has proved to be able to project the vector representation of any word from the source space to the target space accurately. Our vector space projection algorithm is very similar to this.
\section{Pseudo multi-sense detection and elimination by vector space projection}
\subsection{Domain based pseudo multi-sense detection}
\subsubsection{Direct domain similarity}
\par
Given a word and its context, we human beings can easily determine the domains this word belongs to. WordNet makes it convenient for users to get the domains of all synsets of a word. To determine the domain of a sense given the multi-sense word embeddings, we can intuitively define the probability that the $k^{th}$ sense of word $w$ belongs to domain $d$ as 
\begin{equation} \label{pdomain}
P_D(w, k, d) \propto  {\sum_{w' \in NN(w, k)} D(p(w'), d)}
\end{equation}
where $NN(w,k)$ is the nearest neighbors of the $k^{th}$ sense of word $w$ in the given word embeddings, $p(w')$ is the protocol representation of word $w'$ (e.g. when $w'$ is {\sl star\_s1}, $p(w')$ would be {\sl star}), $D(p(w'), d)$ is the sum probability that domain $d$ appears in all synsets of $p(w')$ in WordNet provided by Extended WordNet Domain. Then we can compute the domain similarity between the $k^{th}$ and the $l^{th}$ sense of word $w$ by
\begin{equation}
Sim_D(w,k,l) = \frac 1n|TopN(P_D,w,k,n) \cap TopN(P_D,w,l,n)| 
\end{equation}
where $TopN(P, w, k, n)$ is the set of $x$ that $P(w,k,x)$ ranks top $n$ in decreasing order (in our experiments, $n = 5$).
\subsubsection{Semantic hierarchical similarity}
\par
However, in the knowledge base we applied, the domain knowledge is sometimes not enough for dectecting pseudo multi-sense, especially for some abstract words. For example, it's hard to specify which domain the word {\sl extract} belongs to. What's more, based on \newcite{gonzalez2012graph}, the Extended WordNet Domain cannot reach the precision of 100\%. So we tend to apply semantic hierarchy, particularly hypernymy relations, to help improve our pseudo multi-sense detecting as supplement, since hypernymy somehow contains some domain information. With WordNet, we can also get the semantic relations (e.g. hypernymy, hyponymy, synonymy) of synsets. With the consideration of the DAG structure of semantic relations, for hypernyms of a specific word, the nearer the hypernym, the more information it contains. So we penalize the {\sl far hypernyms}, like {\sl whole, entity, thing}, which cover a large amount of words as their hyponyms. Similar to the definition of $P_D(w,k,t)$, we can define the probability that the $k^{th}$ sense of word $w$ has the hypernym $h$, where $h$ is a synset in WordNet, as
\begin{equation} \label{phyper}
P_H(w, k, h) \propto \frac{1}{d(w,h)} {\sum_{w' \in NN(w, k)} H(p(w'), h)} \cdot \frac{1}{d(p(w'),h)}
\end{equation}
where $d(w,h) = \min_{sw \in Synsets(w)} dis(sw, h)$, $dis(x, y)$ is the distance between two synsets $x$ and $y$ in WordNet, $H(p(w'), h)$ is the frequency that the synset $h$ appears as a hypernym of a synset of $p(w')$ in WordNet. In particular, if $h$ is not a hypernym of $w$ in WordNet, $P_H(w, k, h) = 0$.
\par
We then compute the semantic hierarchical similarity between the $k^{th}$ and the $l^{th}$ sense of word $w$ by
\begin{equation} \label{simh}
Sim_H(w,k,l) = \frac 1n|TopN(P_H,w,k,n) \cap TopN(P_H,w,l,n)| 
\end{equation}
\par
With the definition of domain similarity and semantic hierarchical similarity, we can compute the similarity between the $k^{th}$ and the $l^{th}$ sense of word $w$ by
\begin{equation} \label{simall}
Sim(w,k,l) = Sim_D(w,k,l) + Sim_H(w,k,l)
\end{equation}
\par
When $Sim(w,k,l) > \lambda$, where $\lambda$ is a hyper-parameter ($\lambda = 1$ in our experiments), we consider the $k^{th}$ and the $l^{th}$ sense of word $w$ have the same meaning. In other words, we are able to detect pseudo multi-sense pair $(w_k, w_l)$ based on $Sim(w, k, l)$, which is called pseudo multi-sense detection.
\subsection{Pseudo multi-sense elimination}
\par
Having the existing word embeddings, assume that we have a detected pseudo multi-sense group $G = \{w_{k_1}, w_{k_2}, ... , w_{k_n}\}$, in which $w_{k_1}, w_{k_2}, ... , w_{k_n}$ are senses of word $w$, taking the same meaning. Thus, we can find a representative vector for the group. Let $v_s(w,k_i)$ be the corresponding vectors of $w_{k_i}$, and $v_r(G)$ be the representative vector for the group $G$. Such vector $v_r(G)$ can be randomly chosen from $\{v_s(w,k_1), v_s(w,k_2), ..., v_s(w,k_n)\}$, or simply the mean vector of them. Other methods to compute $v_r(G)$ are also worth trying if reasonable. 
\par
Inspired by \newcite{mikolov2013exploiting}, we assume there is a transition matrix, by which for all pseudo multi-sense group $G$, $\forall w_{k_i} \in G$, $v_{w_{k_i}}$ can be projected to $v_r(G)$. The experiments shown in Section 4 supported our assumption. In other words, we suggest that there exists a global matrix $\Phi$, for any given pseudo multi-sense group $G = \{w_{k_1}, w_{k_2}, ... , w_{k_n}\}$ and its representative vector $v_r(G)$, we have
\begin{equation}
v_r(G) = \Phi * v_s(w, k_i), \forall w_{k_i} \in G, \forall G
\end{equation}
\par
Stochastic gradient descent (SGD) is a stochastic approximation of the gradient descent optimization method for minimizing an objective function written as a sum of differentiable functions by iteration. In order to obtain a consistent $\Phi$ for the projection of all pseudo multi-sense group, we can learn an approximate $\Phi$ with SGD for optimization. Then we  use the obtained $\Phi$ to project existing word embeddings, and thus we can get a new vector space in which pseudo multi-sense has been eliminated compared to the original space.

\section{Experiments}
\par 
We evaluate our pseudo multi-sense detecting and eliminating method both qualitatively and quantitatively. We apply our method to the released word embeddings by \newcite{huang2012improving} and \newcite{neelakantan2015efficient}, which were both trained on the same Wikipedia corpus, and display the performance of our method based on the nearest neighbor task, word similarity tasks and the analogy task. In the following parts, MSSG and NP-MSSG are word embeddings released by \newcite{neelakantan2015efficient}; 50d and 300d are the dimensions of the vector space. The vector space released by \newcite{huang2012improving} are 50-dimensional.

\subsection{Nearest Neighbors}
\begin{table}[htbp]
\centering
\begin{tabular}{|l|l|l|}
\multicolumn{3}{l}{STAR}\\
\hline
\multirow{ 10}{*}{Huang et al.} &
princess, series, cast, serial, midway, sparkle, 1940s, leo, closet, co-star & 01 \\
& silver, boy, cat, version, adventures, stars, emerald, destroyer, terrace, planet& 02\\
& energy, disk, wheel, disadvantage, block, puff, radius, diamond, chord & 03 \\
& version, bronze, standard, colors, ring, emblem, silver, wear, shoulder, red & 01 \\
&workshop, shop, paper, merchandise, plain, corporation, stock, likeness&03\\
&guard, baseball, starter, tennis, basketball, brazil, class, world, morocco, ncaa &01\\
&appearance, entertainer, pat, alumnus, freelance, brother, session, receiver&01\\
&fictional, ongoing, manga, super, japanese, silver, interactive, asian, fiction&01\\
&die, express, ride, opera, spanish, musical, hour, disaster, sun, blue&01\\
&galaxy, spiral, variable, guide, magnitude, companion, satellite, crater&02\\
\hline
\multirow{3}{*}{MSSG-50d} 
& blue, dragon, acbl, diamond, purple, legion, arrow, mercury, eagle, cross & 01\\
& fan, legend, show, moesha, heroes, guest-star, flicka, lassie, tv-movie& 01\\
& stars, sun, constellation, galaxy, eridani, pegasi, supergiant, ceti, starburst&02\\
\hline
\multicolumn{3}{l}{01: person.n.01 ~~~~~~~~~~~~02: celestial\_body.n.01~~~~~~~~~~~~ 03: whole.n.02} \\[0.2cm]
\multicolumn{3}{l}{ROCK}\\
\hline
\multirow{ 10}{*}{Huang et al.} &
blur, indulgence, pop, noise, bands, lacuna, reformed, wave, genre, taster & 01 \\
& energy, silver, cat, song, cd, planet, dawn, hero, video, terrace & 02 \\
& metal, classic, legendary, dubbed, american, hard, belgian, short-lived, debut, da & 01 \\
& soft, shifting, disappear, fill, crystalline, false, pitch, expanse, heat, pile & 03 \\
& vinyl, concert, limited, box, summer, double, dance, enhanced, gold, inch & 04 \\
& hop, well-known, folk, occasional, jazz, music, concert, array, hard, pop & 01\\
& morris, miami, wood, ghost, silver, pearl, chase, corner, oak, thousand & 03 \\
& hard, pop, cm, jazz, hip, hop, r\&b, gutter, wave, subculture & 01\\
& hard, hip, short-lived, classic, jazz, raw, metal, ep & 01\\
& jazz, rally, star, roll, live, entertainer, appearance, session, pop, cover & 01\\
\hline
\multirow{3}{*}{MSSG-50d} 
& metal, rippling, dense, swirling, chirping, blues, punk, psychedelia, bands, pop & 01\\
& sand, rocks, butte, ash, sandy, little, cedar, rocky, sugarloaf, spring-fed& 03\\
& hip, alternative, indie, progressive, hop, reggae, roll, rock/metal, post-hardcore& 01\\
\hline
\multicolumn{3}{l}{01: popular\_music.n.01 ~~~~~~~ 02: person.n.01~~~~~~~~~~~~~~~~03: material.n.01 ~~~~~~~~~~~~~~~~~ 04: whole.n.02} \\[0.2cm]
\multicolumn{3}{l}{NET}\\
\hline
\multirow{ 10}{*}{Huang et al.} &
reduction, amount, increases, stamina, zero, worksheet, improvements, sum & 01 \\
&raw, atomic, destination, brave, orbit, generalize, clock, ca, exhale, fresh & 02\\
&monthly, minimum, retail, banking, dividend, investor, tax, consumer, flat, dollar &03\\
&cash, annual, bribe, yen, generate, yen, liabilities, stocks, lifetime& 03\\
&limousine, panic, alarm, cotton, racket, rush, 9th, buffalo, corps, recovered&04\\
&palm, stalk, blanket, challah, qibla, putting, recess, curtain, tighten, lean&04\\
&indent, text, poser, instruction, libraries, mosaic, campaigns, graphics, imperative&04\\
&freight, processing, volume, needs, passenger, junction, electrical, ferry, shipping&04\\
&contribution, bonus, compensation, bribe, yen, liabilities, stocks, yen, profit&03\\
&1909, quarterback, columbus, bills, bath, elite, 1903, tigers, affiliated, eagles&04\\
\hline
\multirow{3}{*}{MSSG-50d} 
&droplet, pile, wellbore, squeeze, amount, volume, steady, turn, moves, balance&04\\
&boards, run, ball, spot, sideline, at-bat, clock, stretch, running, phils&04\\
&revenue, trillion, assets, profit, billion, pre-tax, liabilities, index, us\$, fdi&03\\
\hline
\multicolumn{3}{l}{01: whole.n.02 ~~~~~~~~~~~~~~~~~~~~~ 02: seize.v.01 ~~~~~~~~~~~~~~~~~ 03: income.n.01 ~~~~~~~~~~~~~~~~~ 04: artifact.n.01}
\end{tabular}
\caption{\label{nntable} Nearest neighbors (by cosine similarity) of sample words and the result of pseudo multi-sense detecting. Column 1 shows the existing word embeddings we use to detect pseudo multi-sense. In Column 2, each row shows the nearest neighbors of one sense in the vector space (Column 1). In Column 3, we present a meaning label for each sense, following the standard of WordNet synset description. We argue that ``senses'' with the same label actually have the same meaning, namely pseudo multi-sense.}
\end{table}
\par
As we hypothesized, previous multi-sense word embedding methods would produce a lot of pseudo multi-sense examples. For the convenience of view, we only focus on the semantic relation in the qualitative evaluation part. We extracted the most probable hypernym for each sense of some sample words by Eq\eqref{simh}, using the synset semantic relations provided by WordNet \cite{miller1995wordnet}. If different representations of one word have the same hypernym, we consider them as pseudo multi-sense. 
\par
In Table~\ref{nntable}, we show the nearest neighbors for each sense of each sample word with multiple word embeddings and our result of pseudo multi-sense detecting. For most of the representations, according to their nearest neighbors, we got reasonable hypernyms. However, there are also some unexpected cases from the result based on the word vectors released by \newcite{huang2012improving}, while no such cases are found in the vectors released by \newcite{neelakantan2015efficient}. For example, we got [whole.n.02] as the hypernym of the three sample words (which seems too general since {\sl whole} can be the hypernym of nearly all entities), and [person.n.01] as a hypernym of {\sl ROCK} (which seems not very reasonable according to the nearest neighbors). By intuition, we suggest that is because of the quality of the word embeddings. Possibly, the level of confidence to extract domains and hypernyms for each sense could be a metric for evaluating the quality of word embeddings. From this point of view, the word embeddings released by \newcite{neelakantan2015efficient} are also with higher quality.
\subsection{Word Similarity}
\par
Now we focus on applying a qualitative evaluation to our method. For each word in the embedded vector space, we first determine the pseudo multi-sense with Eq\eqref{simall}. Then we try to minimize the distance between vectors which belong to the same pseudo multi-sense group, since we argue that they actually represent for the same meaning in the vector space, by training such a matrix $\Phi$, which projects all vectors to a new vector space and eliminate the distance between pseudo multi-sense vectors. We train the matrix $\Phi$ by minimizing the following formula.
\begin{equation} \label{phi}
L = \sum_{(x,x_r)} ||\Phi x- x_r||^2
\end{equation}
where $x$ is a vector which belongs to a pseudo multi-sense group and $x_r$ is the representative vector of the corresponding group. In our experiments, we tried both random sampling and computing mean vector for getting such representative vector.
\subsubsection{Similarity Metrics}
The similarity here is a metric between words to evaluate the performance of word embeddings, which will be used to compare with human judgements, differently from the similarities we introduced in Section 3, which are used to detect pseudo multi-sense.
\par
\newcite{neelakantan2015efficient} introduced three metrics to compute the similarity between words in multi-sense word embeddings, which are $avgSim, avgSimC$ and $localSim$, defined by the following equations.
\begin{equation}
avgSim(w,w') = \frac1K \frac1{K'} \sum_{i=1}^K \sum_{j=1}^{K'} s(v_s(w,i), v_s(w',j))
\end{equation}
where $K$ and $K'$ are the numbers of senses for $w$ and $w'$, $v_s(w,i)$ is the vector of the $i^{th}$ sense of word $w$, and $s(v_s(w,i), v_s(w',j))$ is the similarity measure between vectors $v_s(w,i)$ and $v_s(w',j)$. In our experiments, we apply cosine similarity as $s$.
\par
$AvgSimC$ and $localSim$ can be computed when we have the context of the words.
\begin{equation}
avgSimC(w,w') = \frac1K \frac1{K'} \sum_{i=1}^K \sum_{j=1}^{K'} P(w,c,i)P(w',c',j) s(v_s(w,i), v_s(w',j))
\end{equation}
where $P(w,c,i)$ is the probability for word $w$ to take the $i^{th}$ sense with context vector $c$.
\\
\begin{equation}
localSim(w,w') = s(v_s(w,k), v_s(w',k'))
\end{equation}
where $k = \mathop{\argmax}_i P(w,c,i)$, $k' = \mathop{\argmax}_{i'} P(w',c',i')$.
\subsubsection{WordSim-353}
\par
WordSim-353 is a standard dataset for evaluating the quality of word vectors introduced by \newcite{finkelstein2001placing}, which includes 353 pairs of nouns (without context). Each pair is presented with 13 to 16 human judgements on similarity and relatedness on a scale from 0 to 10. For example, pair (stock, market) gets the score of 8.08, while pair (stock, egg) only gains the score of 1.81. 
\par
In this dataset, since the context of words is not given, we can only compute the $avgSim$ for each pair of word to evaluate our method. The result is shown in Table~\ref{ws353}.
\subsubsection{SCWS}
\par
Stanford Contextual Word Similarity (SCWS) dataset proposed by \newcite{huang2012improving} is also a standard dataset to evaluate the performance of word embeddings quantitatively. It contains 2,003 pairs of words and the context they occur in. 

\par
Then as \newcite{neelakantan2015efficient} did in their work, we also report the Spearman rank correlation between a model's output similarities and the human judgements. We also tried both random sampling and mean vector to get the representative vector for each pseudo multi-sense group. The result of our experiments are shown in Table~\ref{localsimtable}.
 \begin{center}
 \begin{tabular}{|c|ccc|}
 \hline
 \multirow{1}{*}{\textbf{Model}} &\multicolumn{3}{c|}{\textbf{avgSim}} \\
 & original & random & mean\\
 \hline 
 Huang et al. 50d & 64.2& \textbf{65.1} & 65.0\\
 MSSG 50d &  63.2& 65.0 & \textbf{65.1}  \\
 MSSG 300d & \textbf{70.9} & 70.8 & 70.5\\
 NP-MSSG 50d & 62.4 & 64.0 & \textbf{64.4} \\
 NP-MSSG 300d & 68.6 & \textbf{69.1} & 68.8\\
 \hline
 \end{tabular}
 \captionof{table}{\label{ws353} Experimental result on WordSim-353 dataset (Spearman $\rho \times 100$). We apply both random choosing and mean vector to compute the representative vector for each group of pseudo multi-sense. Our method gains a slight improvement on all models except MSSG-300d. }
 \end{center}
 
 \begin{center}
 \begin{tabular}{|c|ccc|ccc|ccc|}
 \hline
 \multirow{1}{*}{\textbf{Model}} &\multicolumn{3}{c}{\textbf{localSim}} & \multicolumn{3}{|c|}{\textbf{avgSim}}  & \multicolumn{3}{c|}{\textbf{avgSimC}} \\
 & original & random & mean & original & random & mean & original & random & mean\\
 \hline 
 Huang et al.  & 26.1 & \textbf{37.6} & 36.9 & 62.8& 61.4 & 62.9 &65.7 & 65.9 & 66.1\\
 MSSG 50d & 49.2 & 52.4 & \textbf{53.2} & 64.2 & 64.9 &  64.8& 66.9 & 67.0 & 67.2\\
 MSSG 300d & 57.3 & 62.1 & \textbf{62.2} & 67.2& 67.3 & 67.2 & 69.3 & 69.1 & 69.4\\
 NPMSSG50d & 50.3 & \textbf{55.5} & 54.9 & 64.0 & 64.1 &64.5 & 66.1 & 66.3& 66.4\\
 NPMSSG300d & 59.8 & \textbf{62.3} & 62.2 & 67.3 & 67.3 & 67.4 &69.1& 68.9& 69.2\\
 \hline
 \end{tabular}
 \captionof{table}{\label{localsimtable} Experimental result on SCWS dataset (Spearman $\rho \times 100$). It shows that the elimination of pseudo multi-sense can significantly improves the performance of word embeddings with the metric {\sl localSim}, while the performances of projected vectors on the metric {\sl avgSim} and {\sl avgSimC} are about the same as those of original vectors. In other words, the elimination of pseudo multi-sense improves the ability of representing a real sense of each sense vector locally.}
 \end{center}
 \subsection{Analogy}
 
 \par 
  Analogy task is another method to evaluate the performance of word embeddings. In single-sense word embeddings, if the word $A$ is similar to word $B$ in the same sense as word $C$ is similar to $D$, there should be an algebraic relationship $v(A)-v(B)=v(C)-v(D)$, where $v(A)$ is the vector of word $A$ in the word embeddings \cite{mikolov2013efficient}. Based on such relationship, we conduct the following experiment, which shows that our method is able to improve the quality of multi-sense word embeddings.
  \par
  In order to compare the quality of different versions of word vectors, our experiment runs on the Semantic-Syntactic Word Relationship dataset, which contains five types of semantic questions and nine types of syntactic questions, as shown in Table~\ref{examples}, including 19544 such quadruples totally.
  
  \par
  For each quadruple in the test dataset, we mark it as $w_1, w_2, w_3, w_4$. The relationship between $w_1$ and $w_2$ is similar to that between $w_3$ and $w_4$. In single-sense word embeddings, we just need to check whether $v(w_4)$ is the most similar vector to $v(w_1)-v(w_2)+v(w_3)$ among all the vectors, and apply the same procedure for $w_1, w_2, w_3$. For multi-sense word embeddings, we check whether there is a combination of senses $\{k_1, k_2, k_3, k_4\}$ so that $v_s(w_4,k_4)$ is the most similar vector to $v_s(w_1, k_1)-v(w_2, k_2)+v(w_3, k_3)$, where $v_s(w,k)$ is the vector of word $w$'s $k^{th}$ sense. What's more, since the equivalence of the two pairs, we also check by such procedure for $v_s(w_1,k_1), v_s(w_2,k_2), v_s(w_3,k_3)$. For every quadruple, once one of the requirements above is satisfied, we treat it as correct. We report the accuracy for each multi-sense vector space in Table~\ref{analogy}.
   \begin{center}
 \begin{tabular}{|c||c|c|c|c|}
 \hline
 Type of relationship & \multicolumn{2}{c|}{Word Pair 1} & \multicolumn{2}{c|}{Word Pair 2} \\
 \hline
 Common capital city & Athens & Greece & Oslo& Norway\\
 All capital cities & Astana&Kazakhstan&Harare&Zimbabwe \\
 Currency & Angola&kwanza&Iran&rial \\
 City-in-state & Chicago&Illinois&Stockton&California \\
 Man-Woman & brother&sister&grandson&granddaughter \\
 \hline
 Adjective to adverb & apparent&apparently&rapid&rapidly \\
 Opposite& possibly&impossibly&ethical&unethical \\
 Comparative &great&greater&tough&tougher \\
 Superlative & easy&easiest&lucky&luckiest \\
 Present Participle & think&thinking&read&reading \\
 Nationality adjective & Switzerland&Swiss&Cambodia&Cambodian \\
 Past tense & walking&walked&swimming&swam \\
 Plural nouns & mouse&mice&dollar&dollars \\
 Plural verbs & work&works&speak&speaks \\
 \hline
 \end{tabular}
 \captionof{table}{\label{examples} Sample quadruple instances in analogy testing dataset. The relations are divided into 5 semantic types and 9 syntactic types.}
 \end{center}
 \begin{center}
 \begin{tabular}{|c|ccc|ccc|}
 \hline
 \multirow{1}{*}{\textbf{Model}} &\multicolumn{3}{c}{\textbf{Semantic}} & \multicolumn{3}{|c|}{\textbf{Syntactic}}  \\
 & original & random & mean & original & random & mean \\
 \hline 
 Huang et al.  & 52.8 & \textbf{53.5} & 53.4 & 53.5 & \textbf{56.1} & 55.9\\
 MSSG 50d & 75.8  & \textbf{77.5}  & 77.4 & 85.2 & 87.9 & \textbf{88.0}\\
 MSSG 300d & 92.0 & 92.8 & \textbf{93.1} & 93.3& 94.1& \textbf{94.5}\\
 NPMSSG 50d & 74.6 & 75.4 & \textbf{75.6} & 80.7 & 82.1 & \textbf{82.3}\\
 NPMSSG 300d & 83.9 & 85.7 & \textbf{85.9} & 89.0 & \textbf{90.2} & 90.1\\
 \hline
 \end{tabular}
 \captionof{table}{\label{analogy} Test result for analogy task. We also apply both random choosing and mean vector to get the representative vector for each pseudo multi-sense group. It shows that our improved vectors perform better on this task.}
 \end{center}
 
 Overall, our detection and elimination of pseudo multi-sense on word embeddings reach higher performance on the nearest neighbor, word similarity and analogy task.
 
\section{Conclusion and future work}

\par
In this paper, we introduced the concept of {\sl pseudo multi-sense}, which is the word embedding models often embed one meaning to multiple senses, to describe the common problem in multi-sense word embeddings. Then we proposed a method based on both domains and semantic relations to detect such cases. What's more, we trained a global transition matrix based on the detected pseudo multi-sense from the given word embeddings, which is used to eliminate the distance between senses actually have the same meaning. The evaluation of our pseudo multi-sense eliminated vector showed that detecting and eliminating pseudo multi-sense significantly improved the ability for each vector in the word embeddings to represent for an exact meaning. We suggest that the following research directions could be considered.% for better performance and deeper understanding of the concept {\sl sense} in multi-sense word embeddings.
\begin{itemize}
\item For the detection of pseudo multi-sense, taking syntactic information and other information we have or we can extract from corpus into account is a reasonable idea to improve the performance.
\item Involve the pseudo multi-sense detection and elimination into the neural network structure, so that the learnt word embeddings could have higher quality than those learnt by existing methods without consideration of pseudo multi-sense.
\item Though we have gained an improvement on experiments, we don't have a deep understanding about the reason that why elimination of pseudo multi-sense works well and why pseudo multi-sense cases are ubiquitous in all kinds of word embeddings. In future work, we could focus on finding a reasonable explanation of the fact.
\end{itemize}


\newpage
\bibliography{coling2016}

\end{document}